\documentclass[wcp]{jmlr}

\usepackage{longtable}

\usepackage{booktabs}

\usepackage{amsmath,amssymb}
\DeclareMathOperator{\E}{\mathbb{E}}

\makeatletter
\let\Ginclude@graphics\@org@Ginclude@graphics 
\makeatother

\jmlrvolume{157}
\jmlryear{2021}
\jmlrworkshop{ACML 2021}

\title[Pedestrian Wind Factor Estimation in Complex Urban Environments]{Pedestrian Wind Factor Estimation \\ in Complex Urban Environments}

   \author{\Name{Sarah Mokhtar} \Email{smokhtar@mit.edu}\\
   \Name{Matthew Beveridge} \Email{mattbev@mit.edu}\\
   \Name{Yumeng Cao} \Email{ymcao@mit.edu}\\
   \Name{Iddo Drori} \Email{idrori@mit.edu}\\
   \addr Massachusetts Institute of Technology, Cambridge, MA 02139, USA}

\editors{Vineeth N Balasubramanian and Ivor Tsang}

\begin{document}

\maketitle

\begin{abstract}
Urban planners and policy makers face the challenge of creating livable and enjoyable cities for larger populations in much denser urban conditions. While the urban microclimate holds a key role in defining the quality of urban spaces today and in the future, the integration of wind microclimate assessment in early urban design and planning processes remains a challenge due to the complexity and high computational expense of computational fluid dynamics (CFD) simulations. This work develops a data-driven workflow for real-time pedestrian wind comfort estimation in complex urban environments which may enable designers, policy makers and city residents to make informed decisions about mobility, health, and energy choices. We use a conditional generative adversarial network (cGAN) architecture to reduce the computational computation while maintaining high confidence levels and interpretability, adequate representation of urban complexity, and suitability for pedestrian comfort estimation. We demonstrate high quality wind field approximations while reducing computation time from days to seconds.
\end{abstract}
\begin{keywords}
Pedestrian Wind Flow Approximation; Conditional Generative Adversarial Networks; Accelerated Computational Fluid Dynamics
\end{keywords}

\section{Introduction}

The global trend towards urbanization, with an expectation of $68\%$ of world’s population living in cities by 2050 \citep{united_nations_2018}, places increased strain on cities. Urban planners, policy makers and built environment professionals face the challenge of creating liveable and enjoyable cities for larger populations in much denser urban conditions. In the face of density, urban outdoor spaces become an increasingly significant contributor to the quality of life, health, and well-being of city residents. Through accommodating daily pedestrian mobility, health, leisure, and social avenues, as well as other various outdoor activity types, urban open spaces contribute to greater physical, environmental, economic, and social benefits for city dwellers \citep{gehl1987life,row1962death}. The quality of those outdoor spaces is highly affected by the outdoor wind microclimate, which in turn is largely defined by the urban morphology \citep{moonen2012urban}. High-rise buildings are responsible for deviating wind at high elevations down to pedestrian levels creating uncomfortable to dangerous conditions. The extensive obstruction characteristic of dense urban centers create large low-wind speed zones which accentuates thermal comfort challenges in hot climates and which, with a warming climate, have the potential to further deteriorate.

While the urban microclimate holds a key role in defining the quality of urban spaces today and in the future, the integration of wind microclimate assessment in early urban design and planning processes remains a challenge. The design process, in practice, is often characterized by uncertainties in design problem formulation as well as frequent and rapid changes, and thus requires a seamlessly integrated simulation workflow \citep{hanna2010beyond}. The complexity and high computational expense associated with computational fluid dynamics (CFD) simulations needed for wind flow estimation make it unsuitable for an iterative design process. Accelerated CFD workflows for urban flow estimation have long been a major area of interest and research in the architectural and urban domains, and remains a bottleneck to the urban design process. This work builds on the opportunities that deep learning workflows offer in modeling highly complex physical phenomena without domain knowledge of the governing physics. By using a conditional generative adversarial network, this work develops a surrogate model which challenges limitations of current implementations, namely constrained representation potential, low confidence, as well as invalidated suitability in the urban micro-climate domain.

\subsection{Related Work}

\paragraph{Deep Learning and CFD:} The field of fluid dynamics deals with enormous amounts of data from field measurements and experiments to larger full flow fields data generated from computational fluid dynamics simulations \citep{brunton2020machine}. This wealth of data, coupled with advances in computing architectures and progress in the field of machine learning in the last decade, have lead to interest in applying deep neural networks (DNNs) for rapidly approximating computational fluid dynamics. Applications of DNNs to fluid dynamics range from physics model augmentation with uncertainty quantification, accuracy prediction improvements, to surrogate modeling for enabling design exploration \citep{kutz2017dlfluiddynamics,uraisamy2019turbulencemodeling}. Convolutional neural networks (CNN) have been particularly explored for the latter due to their capacity to represent non-linear input and output functions while extracting spatial relationships, and generative adversarial networks (GANs) as well due to their additional ability to learn without explicitly defining a loss function. A number of implementations have been successful at reducing the computational expense of velocity fluid flow approximations with a minor error compromise \citep{guo2016,farimani2017deep}. In contrast to other DNN applications such as image and speech recognition, a major challenge in fluid dynamics is the strict requirement for fluid flow fields quantification to be precise, generalizable and interpretable \citep{brunton2020machine}. The computational expense of CFD simulations, additionally, makes it largely unfeasible to repeat experiments and expand datasets. Thus, the finite amount of training data, coupled with distinct feature representation and accuracy requirements across fluid domain disciplines, provides the motivation for the development of application-specific deep learning models, such as this works' focus on pedestrian urban wind flow modeling for early design stages.

\paragraph{Generative Adversarial Networks and their Applications:} In image and video synthesis, frames are typically synthesized using a variational-auto encoder \citep{denton2018stochastic}, auto-regressive model \citep{weissenborn2019scaling}, GAN \citep{vondrick2016generating, saito2017temporal, mirza2014conditional, arjovsky2017wasserstein}, or a combination of these methods. Flow-based methods \citep{kumar2019videoflow} are also utilized in both video and fluid dynamic characterization \citep{ilg2017flownet, babanezhad2020simulation}. In both scenarios, the desired outcome is temporal consistency in the generated fields. Such techniques have been traditionally applied to stylistic tasks \citep{karras2019style}, and have recently made their way into other applications such as driving simulation \citep{kim2021drivegan}. Gaining inspiration from work using GANs to improve wind simulation efficiency \citep{tran2020gans}, we use a conditional generative adversarial network to generate wind comfort estimates in urban environments.

\paragraph{Data-driven Urban Pedestrian Wind Modelling:} Data-driven approaches have been explored intensively in the field of wind modeling and are not recent with experimentation dating back more than to decades \citep{english1999interference}. The focus, in the past, has been on developing workflows that learn to approximate wind flow fields by learning a relationship between an input feature vector typically representing geometry parameters and the corresponding flow field of interest. An instance of an early neural network implementation was to derive wind interference factors from separation distances between two buildings in various wind directions \citep{english1999interference}. Random forest algorithms and other machine learning models were later explored, and training expanded to larger sub-sets of CFD flow fields \citep{wilkinson2013inductive, zhang2004rbf}. Those models were effective in learning relationships between a constrained set of geometrical inputs and rarely provide full fluid flow values. More recent applications enlarge the representation capacity of a surrogate model by integrating more comprehensive urban morphological parameters in the training. Framing this as an optimization problem \citep{wu2021surrogate}, allows finding optimum urban configurations without explicitly estimating flow fields. Deep learning strategies for full velocity fields estimation in urban spaces have also been investigated, including the use of a 3D CNN model \citep{musil2019sustainable}, as well as cGAN models \citep{galanos, mokhtar2020} to create a surrogate that infers wind speed in cities for early design stages. While results are promising, robust evaluation of the suitability of the model for pedestrian comfort applications remains limited, and the confidence levels of the models have not been addressed. The approaches focused on extrusion model representations for buildings, and do not provide an encoding strategy for more complex urban features including canopies, trees and irregular building shapes. 

\subsection{Contribution}
This work challenges the limitations of the deep learning workflows and brings them closer to direct applicability and maximum utility for wind micro-climate applications in order to inform the early architectural design processes. The main contributions of this work are:
\begin{itemize}
\item \textbf{Complex geometry}: Encoding complex building representations beyond simple extrusion geometries and topography,
\item \textbf{Urban elements}: Representing trees and local shading structures,
\item \textbf{Confidence estimates}: Associating wind estimates with model and prediction confidence levels,
\item \textbf{Real time}: Developing a wind surrogate model that reduces the computational time from days to seconds.
\end{itemize}

\section{Methods}
\label{gen_inst}

\subsection{Synthetic Dataset through CFD Simulation}

\begin{figure}
  \centering
  \includegraphics[width=1\linewidth]{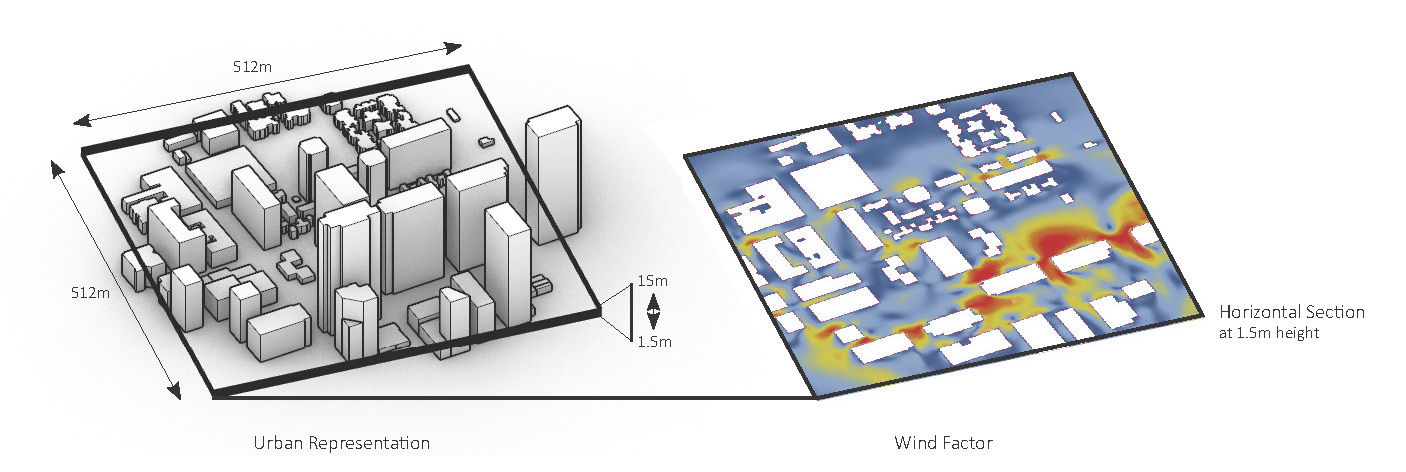}
  \caption{Sample Urban Representation and Corresponding Wind Factors: Representative samples of urban block geometries from a number of cities including San Francisco, Singapore and Cambridge, were selected for the study. For each urban patch, 8 CFD simulations are performed representing wind coming from the eight cardinal directions. For each simulation, 10 horizontal slices of wind factor results are extracted starting at 1.5m above ground at 1.5m intervals. The figure shows one sample of an urban configuration bounded to 512x512 square meters and its corresponding wind factor results.}
  \label{fig:res1}
\end{figure}

We select representative samples of urban block geometries from a number of cities including San Francisco, Singapore and Cambridge, covering a total area of about 3,000 square meters. A diversity in building features, topography, grid typologies and densities is targeted by this selection, using a maximum boundary of 512 square meters. For each urban patch (see Figure \ref{fig:res1}), 8 CFD simulations are performed representing wind coming from the eight cardinal directions. The simulation type and parameters are selected based on industry standards for CFD modeling for pedestrian-level wind applications \citep{franke2007best, windmicroclimate}. The governing equations of steady-state Reynolds-averaged Navier–Stokes equations (RANS) with a realizable k-$\epsilon$ turbulence model are solved using the OpenFOAM solver \citep{openfoam}. The convergence threshold for the simulations is set as wind flow field residuals falling below the recommended $10^{-5}$, which, depending on the geometric complexity, range between $2000-5000$ iterations until termination, and take between $6$ hours to $2$ days of computation running in parallel on a 2.9 GHz i7. A cylindrical domain is used with a radius spanning $15$ times the height of the tallest building, and a height of $6$ times of the tallest building. Across all geometrical configurations, the maximum height ranged between $40-320$ meters. The minimum meshing cell size is defined as $3$ meters with further refinement to the ground, buildings and corners. The inlet wind speed is set as $5$ m/s at a $10$m reference height for all simulations. For each simulation, 10 horizontal slices of the wind tunnel are extracted, in addition to 80 data outputs per urban geometry. A synthetic dataset of input-output pairs are extracted from those simulations in a series of $512 \times 512$ 2D matrices, representing respectively urban geometry and wind factors.

\subsection{Data Encoding}

\begin{figure}
  \centering
  \includegraphics[width=1\linewidth]{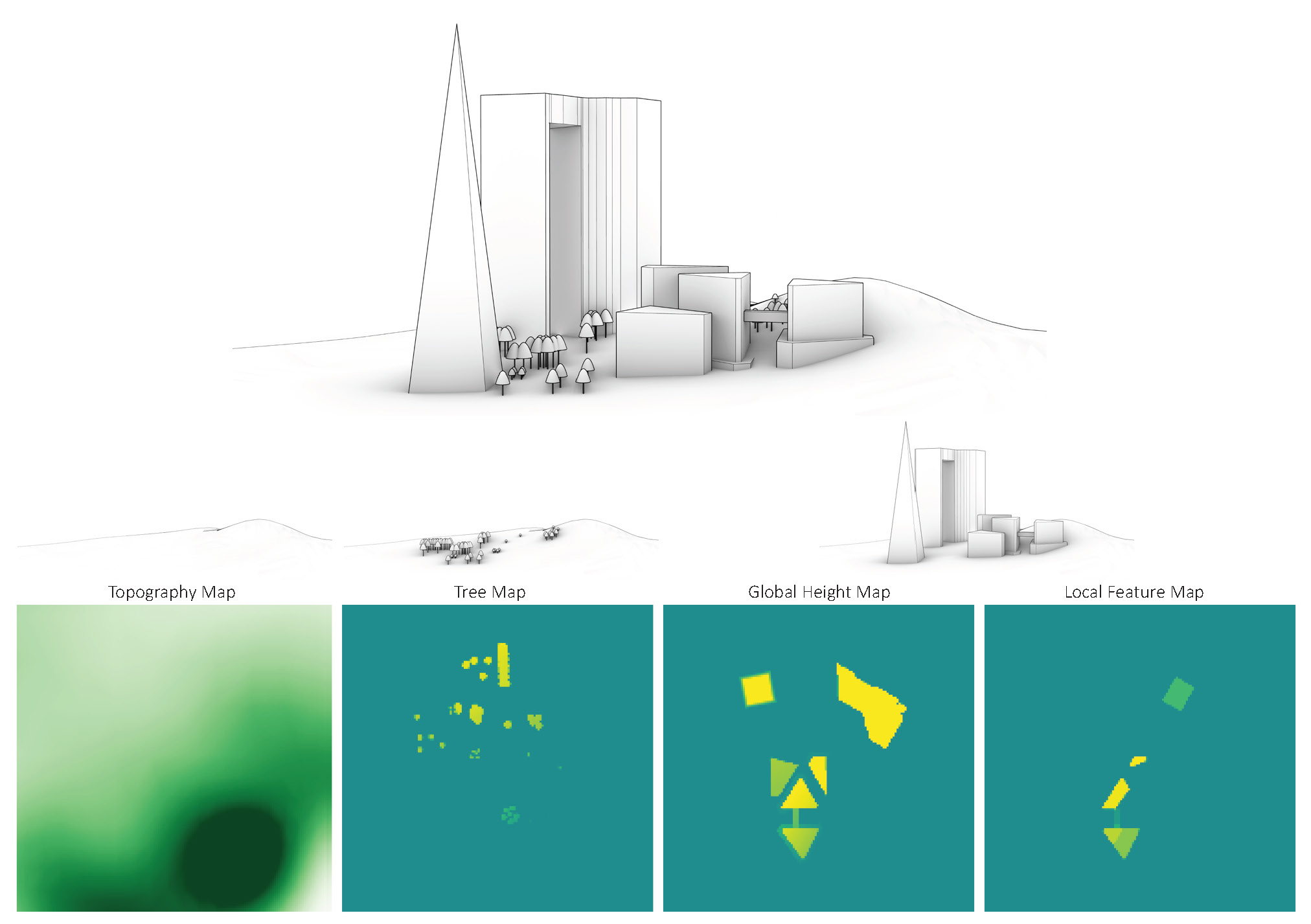}
  \caption{Urban Representation Encoding: Four input channels are used to represent the urban context representing respectively global height features, subtractive local features (such as tunnels, carved balconies, arcades, etc.), topography and additive local features (such as trees, shading devices, etc.).}
  \label{fig:res2}
\end{figure}

\begin{figure}
  \centering
  \includegraphics[width=1\linewidth]{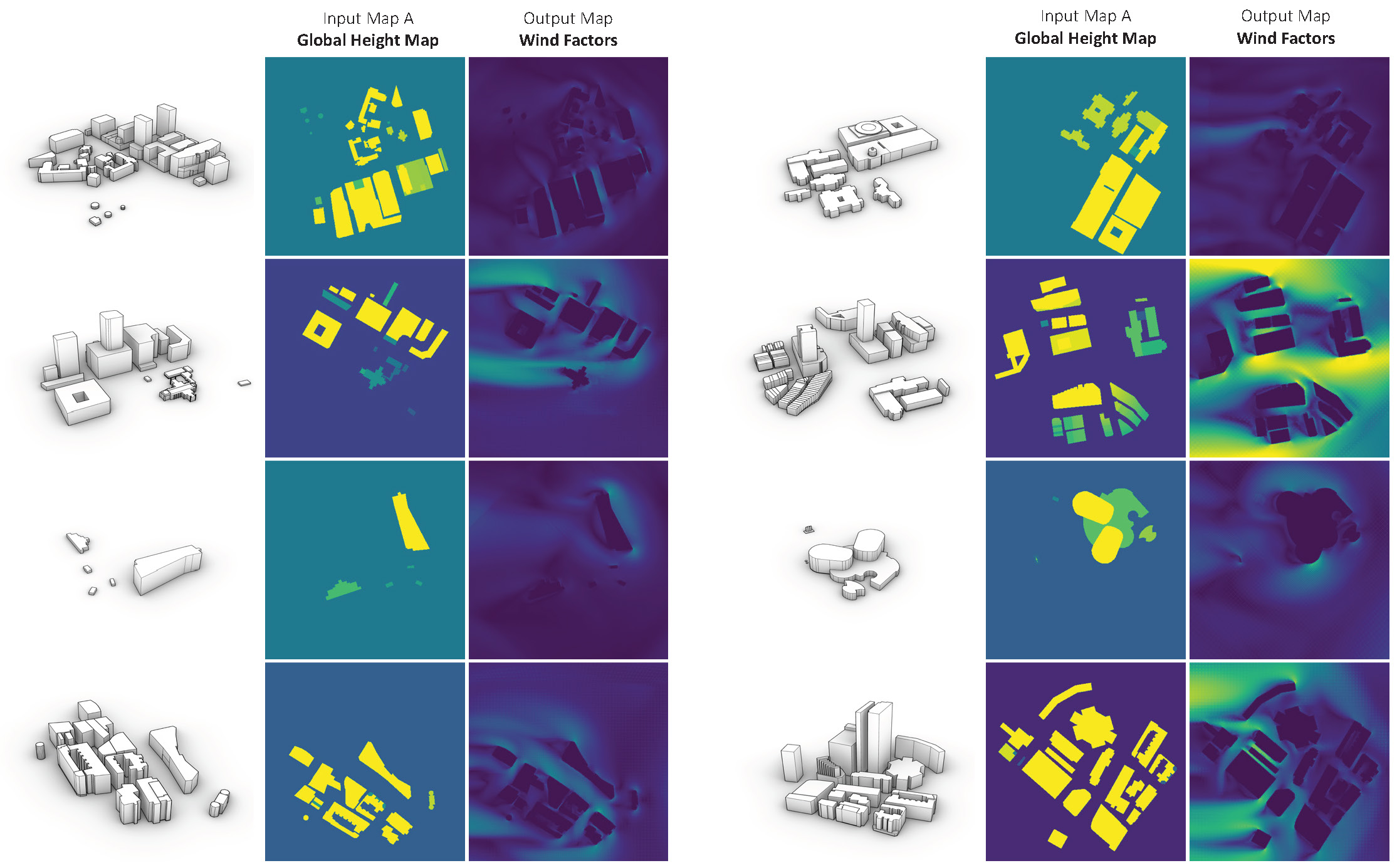}
  \caption{Sample Global Height and Wind Factor Paired Maps: One output channel is defined to each input set and represents wind factor magnitudes. The figure shows a sample of such outputs and their corresponding global height features map and 3D model.}
  \label{fig:res3}
\end{figure}

We take an image-to-image translation approach based on a cGAN architecture. For each simulation, a series of 2D slices of the flow field domain are extracted and associated with 2D matrices representing the urban configuration geometry. Four input channels are defined representing respectively global height features, subtractive local features (such as tunnels, carved balconies, arcades, etc.), topography and additive local features (such as trees, shading devices, etc.), as shown in Figure \ref{fig:res2}. Each consists of a height map on a Cartesian 512x512 grid. The global height map and the tree map are defined such that zero is the height of interest, a negative value represents the distance to the ground and a positive value represents the distance to the furthest obstruction from above. The local feature map extends the representation ability of the global height map by encoding subtractive elements, representing the height of any void above the height of interest. The topography map represents the height above the lowest ground level within the urban boundary. One output channel corresponds to this input set, representing wind factor magnitudes. For each horizontal slice of the wind tunnel, wind speeds are converted to wind factors by normalizing by the simulation inlet wind speed before encoding in a matrix. A sample of global height maps and wind factor magnitude pairs is shown in Figure \ref{fig:res3}.

\subsection{Model Architecture}

An implementation of an image-to-image network architecture \citep{isola2018imagetoimage} is used as our model baseline. The architecture consists of a generator with a Unet skip connection, and a discriminator following the PatchGAN architecture. An additional skip connection block is added to the generator to scale the network to train on larger $512 \times 512$ matrices, and the PatchGAN discriminator output size is increased to a $32 \times 32$ activation map to represent a $140 \times 140$ pixel receptive field. 

\subsection{Uncertainty Estimates using Dropout}
\label{sec:uncertaintyusingdropout}
Dropout is commonly used for neural network regularization during training. In this work, we would also like to learn the posterior over the network weights $p(\theta|x,y) = \frac{p(y|x,\theta)p(\theta)}{p(y|x)}$. However, this in intractable. We therefore estimate uncertainty using dropout. Specifically, we approximate the posterior by sampling using dropout \citep{kendall2017uncertainties}. We train the network using dropout and then test each example $x$ by running multiple forward passes with dropout weights. For $i = 1 \ldots n$ where $n = 30$, we sample $n$ binary masks from a Bernoulli distribution with probability $p$, such that $m_{i} \sim \textit{Ber}(p)$. In this case, we use $p = 0.5$ to generate the Bernoulli masks. We then use $\theta_{i} = \theta \odot m_{i}$, where $\odot$ denotes point-wise multiplication, to compute the mean:
\begin{equation}
    \E(\hat{y}|x) = \frac{1}{n}\sum_{i=1}^{n} f(x|\theta_{i}),
\end{equation}
and use the mean to compute the variance:
\begin{equation}
    v(\hat{y}|x) = \frac{1}{n} \sum_{i = 1}^{n} f(x)^{2} - \E(\hat{y}|x)^{2},
\end{equation}
as an approximation of uncertainty.

\section{Results}

A dataset consisting of 1890 $512 \times 512$ matrices is used for training. To enable a structured assessment of the cGAN predictions, wind factor predictions generated are assessed in comparison to the physically-accurate results from corresponding CFD simulations. The average, standard deviation, and 90th percentile error of the mean absolute error (MAE) between the predicted and simulated wind factors are recorded for all training iterations. This is coupled with a qualitative assessment of generated samples. A disjoint random training and test split of 0.8/0.2 is used, splitting the data into two sets of unique urban configurations. The training is performed for a total of 200 epochs for all iterations, and computation time is 55 seconds per epoch, and 18 seconds per epoch for the resized $25 \times 256$ sets using a GPU. 

\subsection{Model Training}

We train our baseline model using the full $512 \times 512$ px dataset, a 70m receptive field for the discriminator, with a learning rate of $0.0002$ and a generator adversarial to L1 loss ratio of 1 to 100, for 200 epochs. 

\begin{figure}
  \centering
  \includegraphics[width=1\linewidth]{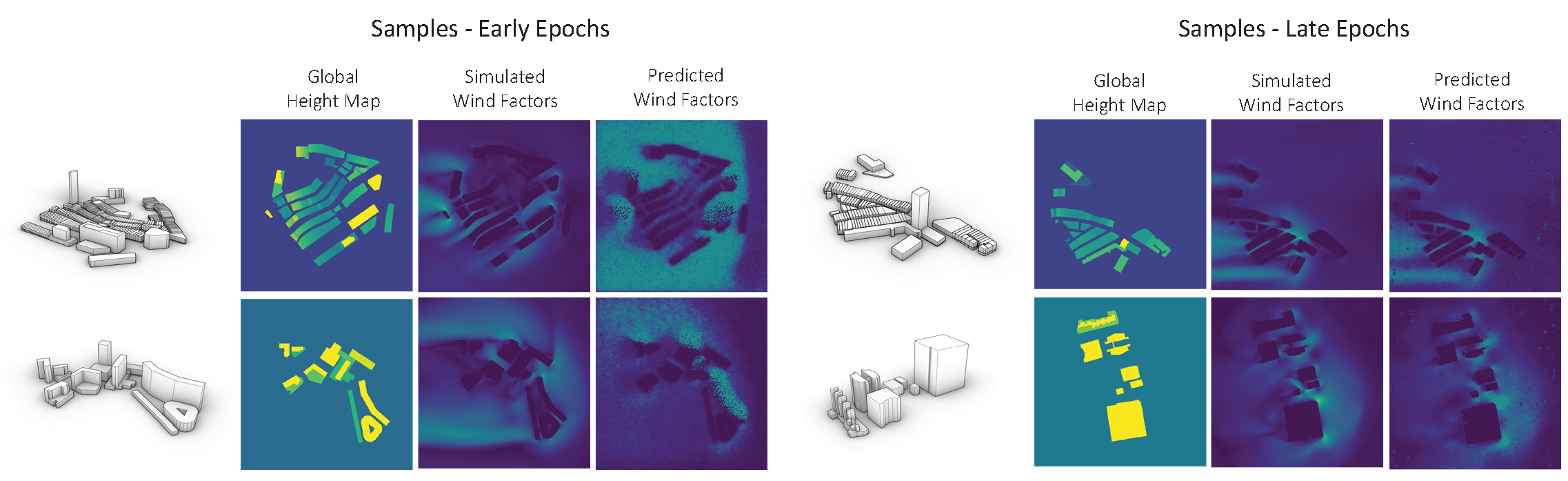}
  \caption{cGAN Predictions as Epochs Increase: Prediction samples are shown at early and late epochs of the training process along with their corresponding ground truth wind factor results, global height map and 3D model. The predictions change from random noise in the early epochs to representations of wind flow identifying zones of influence towards the end of training.}
  \label{fig:res4}
\end{figure}

\begin{figure}
  \centering
  \includegraphics[width=1\linewidth]{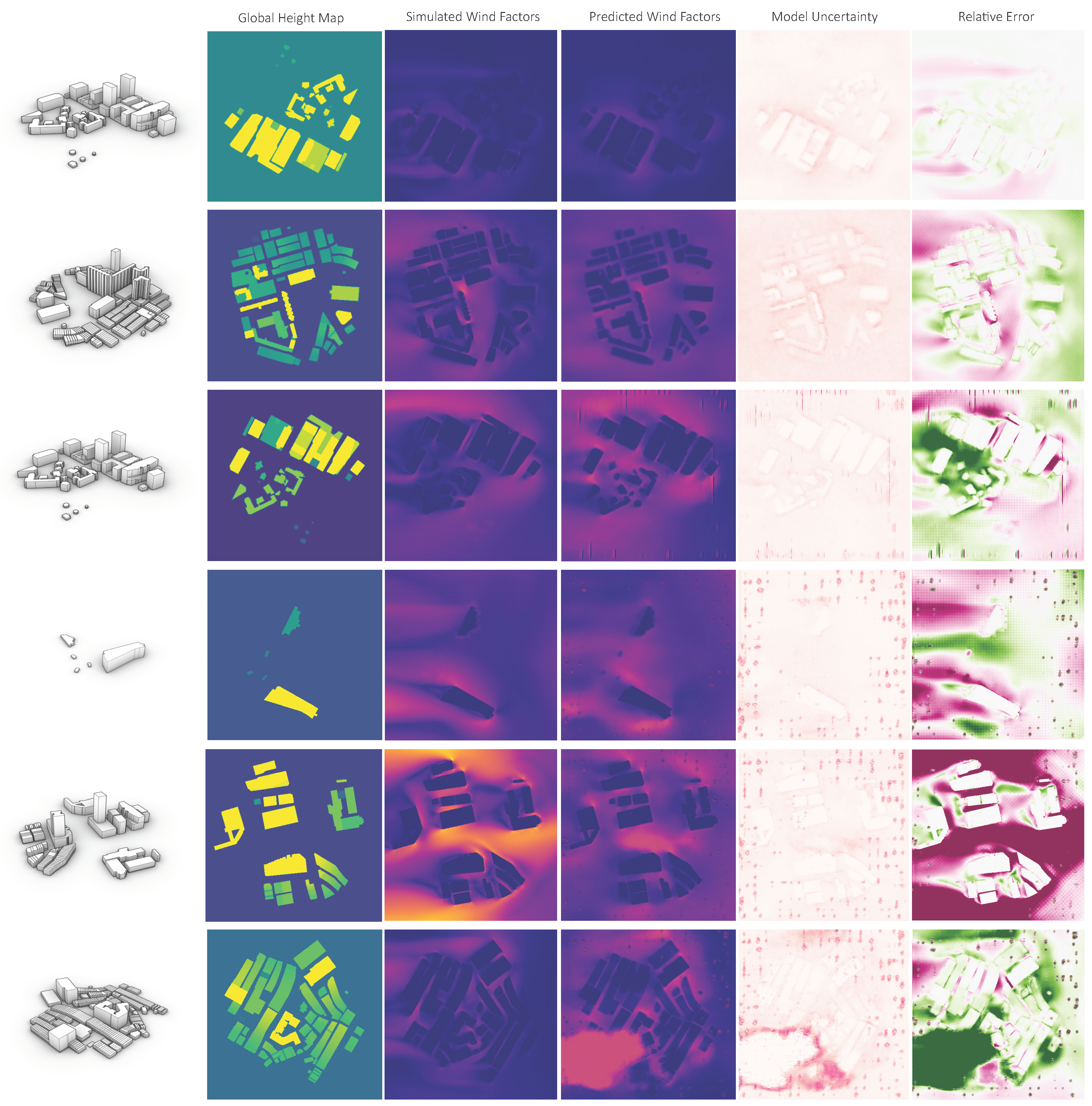}
  \caption{Testing Set Sample Generator Predictions, Uncertainties, and Absolute Errors: A sample of model predictions for select urban patches is shown as well as their associated uncertainties and the absolute error. Visual inspection of results shows the model's capacity to identify zones of impact created by wind obstructions in an urban scene. It also shows its limited capacity to capture the scale of impact for high wind factor zones. Other artifacts include inconsistent color patches in portions of the image, and grainy noise.}
  \label{fig:res5}
\end{figure}

Figure \ref{fig:res4} shows the training progress. The predictions change from random noise in the early epochs to representations of wind flow identifying zones of influence towards the end of training. The mean, standard deviation, and 90th percentile average MAE across all testing pairs are: 0.20, 0.22 and 0.50 respectively. In addition to evaluating the accuracy of model predictions, the model uncertainty is estimated using a Bayesian approximation with dropout as described in Section \ref{sec:uncertaintyusingdropout}. Figure \ref{fig:res5} shows a sample of model predictions, uncertainties, and absolute errors. 

Based on the average MAE recorded and the qualitative assessment of generated output, the cGAN approach is generally capable of identifying the zones of impact created by wind obstructions in the urban scene. A limitation, is that the results are less capable of capturing the scale of impact, and do not capture high wind factor values in around 8\% of the test configurations. Other artifacts which are noticeable in the generated outputs are inconsistent color patches in portions of the image, and grainy noise. Two main factors that contribute to the poorest predictions and artifacts generated are coarse meshing in around 30\% of the CFD simulation outputs, and the data inputs with minor obstructions or obstructions very close to the edges of the image.

\subsection{Implementation Details}
In order to enhance the model prediction accuracy, we tune the input size, receptive field of the discriminator, learning rates of the generator and discriminator, as well as the ratio of adversarial loss to L1 loss for the generator. All iterations are tested as deviations from the baseline model parameters with a $256\times256$ downsized input set size for computational efficiency. Table \ref{tab:enc1} shows the average, standard deviation, and 90th percentile error of the tested configurations.

\begin{table*}
   \small
  \centering
  \begin{tabular}{|ll|c|c|}
    \hline
    \textbf{Parameter} & & \textbf{MAE} & \textbf{90th Perc.}\\
    \hline \hline
    Input Size & \emph{256x256} & $0.19 \pm 0.20$ & $0.45$\\
               & \emph{512x512} & $0.21 \pm 0.22$ & $0.50$\\
     \hline
    Receptive Field & \emph{70} & $0.19 \pm 0.20 $ & $0.45$\\
                    & \emph{284} & $0.18 \pm 0.21$ & $0.44$\\
     \hline
    Learning Rate & \emph{0.0002} & $0.19 \pm 0.20$ & $0.45$\\
                  & \emph{0.001} & $0.19 \pm 0.19$ & $0.46$\\
     \hline
    Adv/L1Loss           &\emph{1/100} & $0.19 \pm 0.20$ & $0.45$  \\
                         &\emph{1/50} & $0.21 \pm 0.24$ & $0.51$  \\
\hline
  \end{tabular}
   \caption{Model Parameters vs. Model Performance: Model parameters are tuned for performance including the input size, receptive field of the discriminator, learning rates of the generator and discriminator, as well as the ratio of adversarial loss to L1 loss for the generator. All iterations are tested as deviations from the baseline model parameters with a $256\times256$ downsized input set size for computational efficiency, and recorded as the mean absolute error, uncertainty and 90th percentile.}
  \label{tab:enc1}
\end{table*}

The results of the hyperparameter tuning process show that increasing the adversarial to L1 loss ratio for the generator results in a reduction in performance, and lower accuracy is observed for larger matrices, with a graceful degradation.

\subsection{Dataset Curation}

Initial experiments show a large impact of the dataset representation and simulation quality on the results generated by the cGAN model. This motivates further explorations into dataset curation. Three different datasets are used in separate training experiments: (i) including only restricted smaller urban contexts, (ii) expanding to include larger contexts and topography, and (iii) which also integrates tree representations. Table \ref{tab:enc2} shows the average, standard deviation, and 90th percentile error of the tested configurations. As expected, the error increases with the complexity of the urban contexts.  

\begin{table*}
  \small
  \centering
  \begin{tabular}{|l|c|c|}
    \hline
    \textbf{Dataset} & \textbf{MAE} & \textbf{90th Perc.}\\
    \hline \hline
    Restricted Context & $0.16 \pm 0.16$ & $0.36$\\
    Extended Context with Topography & $0.19 \pm 0.19$ & $0.45$\\
    Extended Context with Topography and Trees & $0.19 \pm 0.20$ & $0.45$\\
    \hline
  \end{tabular}
  \caption{Dataset Curation vs. Model Performance: Three different datasets are used in separate training experiments: (i) including only restricted smaller urban contexts, (ii) expanding to include larger contexts and topography, and (iii) which also integrates tree representations. The average, standard deviation, and 90th percentile error of the tested configurations are recorded, showing an anticipated increase in error with larger encoded complexities in the urban training set.}
  \label{tab:enc2}
\end{table*}

\subsection{Evaluation and Interpretation}

The evaluation of the model is based on understanding the generator's error ranges with respect to the accuracy required for pedestrian urban wind flow applications. Pedestrian wind comfort criteria, including Lawson, Davenport, and NEN 8100, propose comfort bands that range between $2$m/s for low speeds below $10$m/s and $5$m/s for higher speeds at specific time exceedance thresholds. For inlet wind speeds  at 10m reference height of 2, 3, 4 and 5 m/s, a conservative $\pm$1 m/s error in the prediction would translate to wind factor errors of 0.5, 0.33, 0.25 and 0.2 respectively. The ability of the model to produce high accuracy for urban configurations of various complexities is also relevant to the success of the presented model and thus separate recordings of test errors are captured, as shown in Table \ref{tab:enc3} for the best model.    

\begin{table*}
  \small
  \centering
  \begin{tabular}{|l|c|c|}
    \hline
    \textbf{Urban Complexity} & \textbf{MAE} & \textbf{90th Perc.}\\
    \hline \hline
    Simple Urban Extrusion Models & $0.18 \pm 0.21$ & $0.44$\\
    Urban Models incl. Local Building Features & $ 0.19 \pm 0.22 $ & $0.47$\\
    Urban Models incl. Topography & $ 0.20 \pm 0.22 $ & $0.49$ \\
    Urban Models incl. Trees & $ 0.20 \pm 0.22 $ & $0.49$ \\
    \hline
  \end{tabular}
  \caption{Urban Complexity vs. Model Performance: The trained model's performance in predicting accurately urban wind factor values is tested against urban configurations of various complexities, and testing errors are recorded separately for each set.}
  \label{tab:enc3}
\end{table*}

Based on these error ranges, we achieve an average of 0.2 absolute wind factor error, with minor differences across test data characteristics. The range is within 1 m/s which is a reasonable deviation for cities with wind speeds below 5 m/s, and reflects the merit of our approach. The larger 90th percentile error range, on the other hand, is high, which indicates the potential need for a larger dataset focused particularly on expanding the diversity of configurations, to handle challenging cases. Urban wind flow approximation methods' accuracies in the literature vary significantly. It ranges from reduced-order models (ROM) that can accurately represent turbulent flows but are limited in acceleration rates such as \citep{xiao2019reduced}'s high fidelity non-intrusive ROM achieving 6 times acceleration, and lower accuracy surrogates that achieve substantial accelerations such as the 1500 times faster fast fluid dynamics predictions of \citep{jin2013simulating}. The presented method provides within seconds results of simulations that require at least 6hrs and which can reach 2 days with larger urban complexities.  

In summary, our cGAN approach for estimating wind flow speed for complex urban representations shows promising results with mean error ranges of 0.2. Relative to the intended pedestrian wind comfort application, the results indicate that for the tested urban configurations the errors are within an acceptance threshold of 1 m/s. There are, however, conditions in which the results are not satisfactory for immediate, which may be mediated by using a larger dataset for training and model tuning. Those challenging conditions include urban configurations with two or more high-rise buildings and cities with climates experiencing average wind speeds at 10m reference height above 5m/s. The wind factor predictions of our model, when combined with statistical wind data about any city of interest provides the means to assess wind comfort conditions in urban contexts.

\subsection{Limitations}
Our proposed workflow presents several limitations:
\begin{itemize}
\item Distribution shift: Similar to other machine learning models \citep{kutz2017deep}, ensuring the model's capacity to perform well on test examples that are far outside of training set distribution is challenging. While the dataset used for training is designed to be representative, it does not cover all possible urban morphologies and is created based on a sample selection of city models. As observed in the results, the error ranges may not always be consistent, and a larger repertoire of urban morphologies in the training dataset is thus desirable to improve model usability and confidence predictions.

\item Training set size and resolution scalability: CFD simulations are computationally intensive, with fairly involved post-processing. We identify the lack of large publicly available relevant structured CFD datasets. The expansion of our dataset to include larger urban context and taller buildings would require scaling up computation for performing simulations for generating a larger training set. The quality of the simulation meshing also needs to be fine to prevent the model from learning to replicate artifacts of coarser meshing simulations. As observed in the results, using a lower meshing quality of the simulations included in the dataset may lead to artifacts. This additional constraint makes synthetic dataset expansion more challenging.    

\item Curved surfaces: The encoding strategy implemented in this work challenges existing deep learning urban CFD in the literature by expanding the representation potential to local building features such as shaded entrances, tunnels, bridges, voided structures, carved terraces as well as to topography and trees. While this is in itself an improvement compared with existing applications, it remains limiting in representing curved building elements as well as highly non-uniform vertical obstructions. The exploration of graph-based data-structures as representations of urban configurations may provide additional representation power in this context.        

\end{itemize}

\section{Conclusions}

We present a methodology for surrogate modeling of wind flow approximation in urban contexts, improving upon the current implementations by representing building local features, topography, and trees. The results show the model's capacity to infer wind factors at an accuracy of $\pm 0.2$ within seconds. Compared to an equivalent CFD simulations, the predictions are at a fraction of the time and at an accuracy expense that is acceptable for pedestrian wind comfort. Further work is needed to expand the applicability and usability of the model to larger, more complex and taller urban conditions. Dataset expansion is needed to enable the model to learn from a larger repertoire of urban contexts.

Quantifying the urban microclimate is an essential step towards enabling urban planners, policy makers, and built environment professionals to make informed decisions about the design of liveable cities that are comfortable and resilient to climate change. Our approach brings this vision closer to larger usability in the design of cities by providing a fast approximation for estimate wind speeds at pedestrian levels at high quality. We make our code available in the supplementary material, removing barriers to entry, reaching beyond domain experts in CFD simulations.

\bibliography{acml21}

\end{document}